\documentclass{article}
\usepackage[affil-it]{authblk}
\usepackage{color}
\usepackage{amsmath,amssymb}
\usepackage{tikz}
\usepackage{pgflibraryarrows}
\usepackage{pgflibrarysnakes}
\usetikzlibrary{arrows,shapes}
\usetikzlibrary{positioning,calc}
\usepackage{float}
\usepackage[natbibapa]{apacite}
\usepackage{hyperref}
\usepackage{makeidx} 
\usepackage{comment}
\usepackage{indentfirst}
\usepackage{subfigure}
\usepackage{geometry}
\usepackage{epstopdf}
\usepackage{hyperref}
\usepackage{color}
\usepackage[normalem]{ulem}
\newcounter{ToDo}
\newcounter{gaocomm}
\newcounter{Note}
\definecolor{blue-violet}{rgb}{0.54, 0.17, 0.89}
\definecolor{mygreen}{rgb}{0.0, 0.5, 0.0}
\definecolor{awesome}{rgb}{1.0, 0.13, 0.32}
\definecolor{bostonuniversityred}{rgb}{0.8, 0.0, 0.0}

\begin{document}

\title{Assessing the Performance of Deep Learning Algorithms for Newsvendor Problem}

\author{Yanfei Zhang%
	\thanks{Email address: \texttt{yzha4636@uni.sydney.edu.au}}}
\affil{The University of Sydney Business School\\
	The University of Sydney\\
	Sydney, NSW 2006, Australia\\}

\author{Junbin Gao%
	\thanks{Email address: \texttt{junbin.gao@sydney.edu.au}}}
\affil{The University of Sydney Business School\\
	The University of Sydney\\
	Sydney, NSW 2006, Australia\\}
 
\maketitle
\begin{abstract}
In retailer management, the Newsvendor problem has widely attracted attention as one of basic inventory models. In the traditional approach to solving this problem, it relies on the probability distribution of the demand. In theory, if the probability distribution is known, the problem can be considered as fully solved. However, in any real world scenario, it is almost impossible to even approximate or estimate a better probability distribution for the demand. In recent years, researchers start adopting machine learning approach to learn a demand prediction model by using other feature information. In this paper, we propose a supervised learning that optimizes the demand quantities for products based on feature information. We demonstrate that the original Newsvendor loss function as the training objective outperforms the recently suggested quadratic loss function. The new algorithm has been assessed on both the synthetic data and real-world data, demonstrating better performance.  
\end{abstract}

\section{Introduction}
Two recent papers \citep{RudinVahn2013} and \citep{OroojlooyjadidSnyderTakac2016} discuss the machine learning approach for the classical Newsvendor problem. The classical Newsvendor problem optimizes the inventory of a perishable good under the assumption that the probability distribution of the demand is fully known. Perishable goods are those that have a limited selling season. A retailer may order or purchase the goods at the beginning of a time period and sell them during the period. For whatever reasons, after certain time or at the end of the period, the retailer must dispose of unsold goods. This cause the so-called holding cost. On the other hand, if the good is highly demanded in the period, the retailer may soon run out of the goods, thus it incurs a opportunity cost resulting from the shortage (denoted by 'shortage cost'), resulting in potential profit loss. Hence for the best profit, the optimal order quantity for the Newsvendor problem should be sought to minimise the expected sum of the two costs for the retailer.  

The above problem can be formulated as an optimisation problem as follows:
\begin{align}
	\min_y\quad C(y)=E_d[c_p(d-y)_{+}+c_h(y-d)_{+}], \label{Eq1}
\end{align} 
where $d$ is the unknown demand, $y$ is the order quantity, $c_p$ and $c_h$ are the per-unit shortage and holding costs, respectively, and $(a)_+:=\max\left\{0,a\right\}$. This objective function is called as 'original loss function' in this article.

The classical solution assumes that the demand follows some underlying distribution, for example, a normal distribution. Under that assumption, the optimal order amount can be solved as, see \cite{GallegoMoon1993},
$
	y^*=F^{-1}\left(\frac{c_p}{c_p+c_h}\right).
$

The obvious hurdle to apply this approach is how to get the demand distribution. Also the one-product nature for this original loss function \eqref{Eq1} is also a problem for empirical use. As the distributional information usually can be in strong assumptions, which are most likely unknown in real life, relying on the distributional information is not a plausible method. Thus developing a new model to make it independent from too many strong assumptions is quite important. 

Recently, Inspiring by the 'big-data', some of the researchers tried to use machine learning approaches (especially Deep Learning) to solve the distribution-free version of the problem. However, the original loss function is non-differentiable, making the general back-propagation (BP) algorithm in machine learning unfeasible. Thus, this article focuses on the following problems:

\begin{enumerate}
\item We demonstrate that the original loss function in \eqref{Eq1} can be integrated into any neural network architecture and the neural network training can run smoothly; 
\item We test whether original loss function is indeed comparable or superior to using the Quadratic loss function, first suggested by \cite{OroojlooyjadidSnyderTakac2016}; and
\item We analyse the influence of both $c_p$ and $c_h$ in Deep Learning neural network training.
\end{enumerate}

The paper is organized as follows. In Section \ref{Sec:2}, we summarise the major literature in newsvendor problem research. Section \ref{Sec:3} focuses on expressing the related works and introduce the basic machine learning setting for the Newsvendor problems. introducing the Deep Learning neural network architecture and derive the BP algorithm when the proposed L1 loss function is integrated.  In Section \ref{Sec:4}, the performance of the proposed method is evaluated on both data and real-world datasets. Finally, conclusions and suggestions for future work are provided in Section \ref{Sec:5}.

\section{Previous Works} \label{Sec:2}

Early research mainly focuses on the refinement of distributional and mathematical method, and solving the model as an optimisation problem. For example, \cite{LauLau1988} designed an algorithm for the price-dependent distribution method to exclude the influence of price. The multi-product Newsvendor under assumed demand distribution was also considered by some researchers.\cite{ZhouChenXuYu2015} proposed a method for the extension of the distribution method to multi-product cases. Some researchers consider the previous Newsvendor model with distributional assumption in terms of multi-period. For example, \cite{Alwan2016} considered the problem when the demands from different periods have correlation and would cause effect to the subsequent period. They applied the AR(1) model on the Red Blood Cell data from an American Regional Hospital. However, the outcome shows that the correlation in different period has no effect on the prediction. Besides, \cite{ShuklaJharkharia2011,BoxJenkinsReinselLjung2015} also proposed similar methods.

As the distributional information usually can be a strong assumption, which is mostly unknown in real life, relying on the distributional information is not a plausible method. Thus developing a new model to make it independent from many assumptions is quite important. \cite{Scarf1957} first tried to solve the Newsvendor problem with only sample mean $\bar{x}$ and sample variance $\hat{\sigma}^2$ given (instead the whole distribution information). Motivated by \cite{Scarf1957}, \cite{GallegoMoon1993} further expand Scarf's model to multi-product case by calculating the demand for each item and simply add them up. However, at this stage, they hadn't integrated the effect of data features into the analysis, thus generate a biased outcome.\cite{RudinVahn2013} further tried to solve the multi-product Newsvendor problem in a more plausible way, by assuming the optimal order quantity as the affine function of data features. 

Recently, Newsvendor problem is encouraged by the concept of big-data. The earliest work can be seen in \cite{CarbonneauLaframboiseVahidov2008} where the classic neural network and recurrent neural networks techniques were applied to the demand/order time series. \cite{ShiChenWangYeungWongWoo2015} had proposed a LSTM neural network approach in solving Newsvendor-like weather precipitation nowcasting. The previous two researches predicted optimal distribution, rather than directly the order amount, which is sub-optimal. 

Under these circumstances, \cite{OroojlooyjadidSnyderTakac2016} improved the previous method by incorporating both the method from \cite{ShiChenWangYeungWongWoo2015} and \cite{RudinVahn2013}. To avoid the non-differentiable original loss function \eqref{Eq1}, a Quadratic loss function was proposed to derive the gradient for the implementation of back-propagation algorithm in training neural networks. However, it is well-known that the Quadratic loss function may cause an over-fitting problem which might cause distortion due to the existence of outlier in the training data. In machine learning research, a more appropriate loss function against outliers is the so-called 'L1-norm loss function', i.e., the original loss function \eqref{Eq1} mentioned previous in this paper, see \cite{Bishop2006}.

\section{Methodology And Major Theoretical Contribution} \label{Sec:3}

\subsection{Machine Learning Setting for Newsvendor Problems}
In this subsection, we present the details of machine learning setting for classic Newsvendor problems. We assume that $N$ historical observations are available, which are denoted as $\{\mathbf{d}_i\}^N_{i=1}$ where each $\mathbf{d}_i\in\mathbb{R}^m$ is a vector of demand information for $m$ goods. A number of $p$ observable features is attached to each vector of demand data $\mathbf{d}_i$, collected in a vector in dimension $p$ as $\mathbf{x}_i\in\mathbb{R}^p$. The full set of observed data consist of $N$ set of features and demand, that is
$
\mathcal{D}_N = \{(\mathbf{x}_i, \mathbf{d}_i)\}^N_{i=1}.
$

For the given dataset $\mathcal{D}_N$, the machine learning task is to learning a mapping $f$ from the feature vector $\mathbf{x}\in\mathbb{R}^n$ to the demand vector $\mathbf{d}\in\mathbb{R}^m$ under certain criterion.

Considering the original loss function defined in \eqref{Eq1}, the most appropriate specification under the context multi-product Newsvendor model is
\begin{align}
\min_f \sum^N_{i=1}\left\| c_h(\mathbf{d}_i-f(\mathbf{x}_i))_+ + c_p(f(\mathbf{x}_i) - \mathbf{d}_i )_+\right\|_1 \label{Eq2}
\end{align}
where $(\cdot)_+$ operator operates on each component of the vector and $\|\cdot\|_1$ means the L1-norm, i.e., the sum of absolute values of components of an $m$-dimensional vector\footnote{Note: Both $c_h$ and $c_p$ can be defined individually for each demanded product/goods.}.

\subsection{Related Previous Works}
To complete the machine learning setting in \eqref{Eq2}, we shall specify the model space for the mapping $f$. There are plenty of choices for this purpose. \cite{RudinVahn2013} solves the multi-product Newsvendor problem by assuming the optimal order quantity was linear combination of the features adjusted by parameters. To be more specific, the mapping $f$ is defined as 
\begin{align}
    f(\mathbf{x}_i)=q_0+\mathbf{q}^T \mathbf{x}_i=q_0+\sum_{j=1}^{n}q_{j} x_{ij},
    \label{Eq3}
\end{align}
where $\mathbf{q}=[ q_1,\cdots ,q_n]^T \in\mathbb{R}^n$, a set of weights that is to be fitted to data $\mathbf x_i$'s, and $q_0$ is a disturbance ('intercept') term, these parameters minimises \eqref{Eq2}.

Further more,
\cite{OroojlooyjadidSnyderTakac2016} combined the formulation of \cite{RudinVahn2013} and the Deep Learning Neural Network (DNN) to better capture nonlinear relation between data features and demand quantity. A DNN with 2 hidden layers and Sigmoid activation functions was introduced. This way has parameterised the mapping $f$ in terms of DNN which defines a highly nonlinear mapping. To avoid the non-differentiable objective function in \eqref{Eq2}, they instead use a L2-norm loss function to derive the gradient to better implement the BP algorithm for all the network weights, which can be written out as:
\begin{align}
\min_{\mathbf{q}} \sum^N_{i=1}\left\| c_h(\mathbf{d}_i-f(\mathbf{x}_i, \mathbf{q}))_+ + c_p(f(\mathbf{x}_i, \mathbf{q}) - \mathbf{d}_i )_+\right\|^2_2, \label{Eq4}
\end{align}
where the notation $\mathbf{q}$ collects all the neural network weights and $\|\cdot\|_2$ is the L2 norm. We call this loss function 'Quadratic loss function' in comparison with the original loss function \eqref{Eq2},

\textit{Remark 1:} Although the objective function defined in \eqref{Eq4} has been successful as reported in \cite{OroojlooyjadidSnyderTakac2016}, we argue that the proposed loss function is indeed inappropriate for the newsvendor problems. To see this, assume that all the inputs $\mathbf x_i$ are all the same, then the learning problem goes back to the classical newsvendor problems with distribution free approach where we simply predict a single demand quantity. In that case, the objective \eqref{Eq4} is clearly different from the empirical form of \eqref{Eq1}. Our proposed lost function \eqref{Eq2} becomes \eqref{Eq1} (its empirical mean). Our experiments also demonstrate the effectiveness of \eqref{Eq2}.

\textit{Remark 2:} Nothing prevents us from using other models for the above machine learning task for the newsvendor problems, for example, the mapping $f$ can be defined by a Gaussian Process \citep{Bishop2006} or other data-driven kernel models \citep{GaoKwanShi2010,GaoGunnHarris2003a}.

\textit{Remark 3:} For convenience, we will denote all the features in a matrix $\mathbf X\in\mathbb{R}^{N\times n}$ and the demand quantity in $\mathbf{D}\in\mathbb{R}^{N\times m}$. For example, for 2 products and for past 2-week length of time, the product has 14 historical demands, which is an $14\times 2$ matrix $\mathbf D$. Each historical demand is with 3 data features: holiday (1 for holiday), weather (1 for bad weather), promotion (1 for promotion of today). The total data features should be a $14\times 3$ matrix.

\subsection{Theoretical Contribution}
As in \cite{OroojlooyjadidSnyderTakac2016}, this paper will consider a mapping $f(\mathbf{x}_i, \mathbf{q})$ defined by a classic DNN under the original loss function \eqref{Eq2}, which was non-differentiable at some points. To derive the BP algorithm for the neural network modeling based on the new objective \eqref{Eq2}, we will top up one more layer on the output $f(\mathbf{x}, \mathbf{q})$ from the classic neural networks. 

It is easy to see that the objective function in \eqref{Eq1} can be decomposed into two ReLU (Rectified Linear Unit) units \citep{HahnloserSarpeshkarMahowaldDouglasSeung2000}, as shown in Fig.\ref{NN}. In fact, this comes from the fact that each term in the loss function of \eqref{Eq1} can be written as the following form:  
$$ C(f(\mathbf x_i,\mathbf{q}))=
 \begin{cases}
c_p \max(\mathbf{d}_i-f(\mathbf{x}_i,\mathbf{q}),0),& \mathbf{d}_i\geq f(\mathbf{x}_i,\mathbf{q}) \\ 
c_p \max(f(\mathbf{x}_i,\mathbf{q})-\mathbf{d}_i,0),& f(\mathbf{x}_i,\mathbf{q})\geq \mathbf{d}_i  
\end{cases} 
$$  
which is coincidently same as two ReLUs.  

\tikzstyle{format} = [draw, thin, fill=blue!20]
\begin{figure}[H]
\begin{center}
\begin{tikzpicture}[node distance=1cm, auto,>=latex', thick]
\path[->]node[format](NNInput) {$f(\mathbf{x}, \mathbf{q})$ from NN};
\path[->]node[format, right= 2cm of NNInput](D) {$\mathbf d$ Target Demand} ;
\path[->]node[format, above= 1cm of {$(NNInput)!0.3!(D)$} ] (ReLU1) {ReLU 1} (NNInput) edge node {$-c_p$} (ReLU1)(D) edge node {$c_p$} (ReLU1);
\path[->]node[format, above= 1cm of {$(NNInput)!0.7!(D)$} ] (ReLU2) {ReLU 2} (NNInput) edge[below] node {$c_h$} (ReLU2)(D) edge[right] node {$-c_h$} (ReLU2);
\path[->]node[format, above= 1cm of {$(ReLU1)!0.5!(ReLU2)$} ] (Plus) {$+$} (ReLU1) edge node{} (Plus) (ReLU2) edge node{} (Plus) ;
\path[->]node[above= 1cm of Plus ] (L) {Loss}  (Plus) edge node{} (L);
\path[->]node[format, below= 1cm of NNInput ] (NN) {Classic NNs Accepting $\mathbf x$} (NN) edge node{} (NNInput);
\end{tikzpicture}\caption{The proposed neural network structure for $m=1$.}\label{NN}
\end{center}
\end{figure}
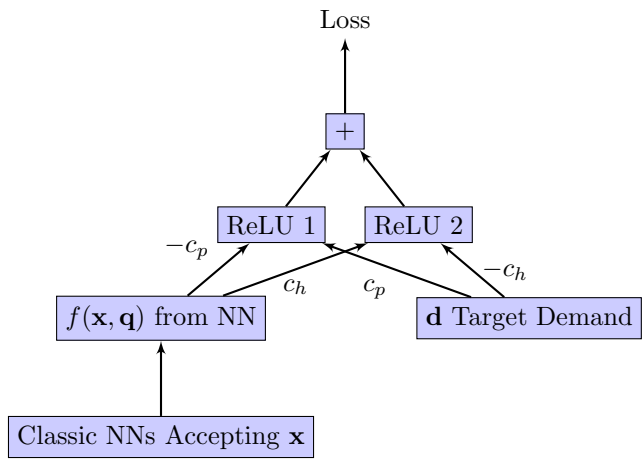 

The reason why this kind of structure can be successfully implemented in a BP algorithm is that, although at 0 both ReLUs are non-differentiable, it is quite rare to happen in real life such that $\mathbf{d}_i$ equals to $f(\mathbf{x}_i, \mathbf{q})$. The successful application of ReLUs in the state-of-the-art Deep Learning architecture has demonstrated this. Thus, the gradient of the loss function can be expressed as:
			$$ \nabla_{f(\mathbf{x}_i, \mathbf{q})} C(f(\mathbf{x}_i,\mathbf{q}))=
				\begin{cases}
				-c_p &\mathbf{d}_i\geq f(\mathbf{x}_i,\mathbf{q}), \\
				c_h & f(\mathbf{x}_i,\mathbf{q}) > \mathbf{d}_i.  
				\end{cases}
			$$
Thus, the gradient function for the original loss function can be obtained:
\begin{align*}
\frac{\partial C(f(\mathbf{x}_i,\mathbf{q}))}{\partial f(\mathbf{x}_i,\mathbf{q})}
=-c_p\delta (\mathbf{d}_i>f(\mathbf{x}_i,\mathbf{q}))+c_h\delta(f(\mathbf{x}_i,\mathbf{q})>\mathbf{d}_i),
\end{align*}
where $\delta$ is the condition indicator function such that $\delta(\text{true})=1$ and $\delta(\text{false})=0$.  

Subsequently, the gradient for the subsequent weights can be decomposed like what had been done in \cite{OroojlooyjadidSnyderTakac2016}, and the gradient descent algorithm can be applied on finding the optimised weights for each path to generate the smallest Newsvendor cost. In the next section we will follow the standard machine learning protocol to conduct modeling training and model testing to empirically demonstrate our claim.

As a strategy of common practice in neural network training, we also add the following quadratic regularisation on the neural networks weights $q$ to the objectives \eqref{Eq2} and \eqref{Eq3}, respectively,
$
R(\mathbf{q}) = \lambda \sum_{i,j}q^2_{ij},
$
where $\lambda>0$ is the regularisation term to trade-off between cost and magnitude of weights. In our experiments we find that the training was not highly influenced by the value of $\lambda$, so we set $\lambda=10^{-3}$. 

\section{Numerical Experiment} \label{Sec:4}
As previously mentioned, the major computation of the numerical experiment is undertaking by a Deep Learning Neural Network (DNN). As the following numerical experiments are inspired by the research in \cite{OroojlooyjadidSnyderTakac2016}, the structure of neural network in this paper would be similar to the 2-hidden-layer DNN for a fair comparison, but the specific parameter setting (number of neurons in 2 hidden layers, the regularisation term $\lambda$ and scaling parameter $f$) would not be the same. 

\subsection{Experimental Setting and Performance Assessment Criteria}
In general, we will split the given dataset into two parts: one for training to generate optimised model parameters $\mathbf{q}$, in abbreviation, we call it 'training set'. The rest of the data is generally used for testing the performance of the specific prediction method, which is called 'testing set'. 

For our convenience, we denote the training and testing sets, respectively, as
\[
\mathcal{D}_{\text{train}}=\{(\mathbf{x}^{\text{train}}_j, \mathbf{d}^{\text{train}}_j)\}^{n_{\text{train}}}_{j=1}
\text{ and }  
\mathcal{D}_{\text{test}}=\{(\mathbf{x}^{\text{test}}_j, \mathbf{d}^{\text{test}}_j)\}^{n_{\text{test}}}_{j=1}.
\]

Accordingly, to assess the performance of two loss functions, as usual, we propose to use the following training error and 
testing error, as defined respectively by,
\begin{align}
	\text{TestErr} &=\frac{1}{n_{\text{test}}}\sum_{j=1}^{n_{\text{test}}}\|\hat{f}(\mathbf{x}^{\text{test}}_j)-\mathbf{d}^{\text{test}}_j\|_2^2, \label{Eq5}\\
    \text{TrainErr} & =\frac{1}{n_{\text{train}}}\sum_{j=1}^{n_{\text{train}}}\|\hat{f}(\mathbf{x}^{\text{train}}_j)-\mathbf{d}^{\text{train}}_j\|_2^2.
\end{align}
where $\hat{f}(\cdot)$ is the predicted demand from the model testing data $\mathbf{x}^{test}_i$, 
and $\mathbf{d}_i$ is the demand for each observation in the training/testing set and $\|\cdot\|_2$ is the L2-norm in $\mathbb{R}^m$. The training error can be used for cross validation.This method tests the variation for the error term, smaller testing error indicates small variation in error term, i.e. good fitness.

Similarly the following training error can be used for model The second one used for cross validation, defined as follows,
\begin{align}
 	\text{TrainErr}=\frac{1}{n_{\text{train}}}\sum_{j=1}^{n_{\text{train}}}\|\hat{f}(\mathbf{x}^{\text{train}}_j)-\mathbf{d}^{\text{train}}_j\|_2^2. 
 	\label{Eq6}
\end{align}

As previously illustrated, our objective is to find out whether original loss function would be better in terms of overfitting problem. If the model have overfitted the problem, its predictability (assessing by testing set) would be poor, while in-sample (training set) fitness can be small. As these two selection criteria both measure the variation in the error term, if the model have a good predictability, large training errors and small testing errors are expected in the following experiments.

Each training error and testing error would be displayed against $c_p/c_h$, the ratio of shortage costs over holding costs. Decomposing both loss functions, $c_p$ and $c_h$ determines the magnitude of loss function, and we believe that the $\mathbf{q}$ would be affected in the minimisation process. $c_p/c_h$ from 1 to 10 with the step length of 0.5 is introduced to better capture the change between two integers. 

\textit{Remark 1:} An interesting truth we found in numerical experiment was that, $c_h$ cannot be setting to 1, otherwise the prediction for original loss function would just fluctuates in a small range, which cannot fully recover the predictability of it. Thus, the value for $c_h$ was set as 1.5 and $c_p$ as $c_h\times 1$ to $c_h\times 10$.

The algorithm in this paper was implemented by using Mathworks MATLAB 2017a and all the experiments were conducted on a laptop with a CPU Intel i7-6500U and an memory size of 8GB. The BP algorithm is implemented by the L-BFGS procedure due to its less memory usage property, comparing to Quasi-Newton and BFGS methods.

\subsection{Experiments on the Synthetic Dataset}
To quickly assess the performance of the Newsvendor objective function in \eqref{Eq2}, we conduct a numerical experiment on an small synthetic dataset. This dataset is with features that is consisted by three binary variables for the Weather condition, Holiday, and Promotion. There are two weeks demands data for both training and testing respectively (a total of 28 observations, half for training and half for testing).

The architecture of the neural networks was configured in the following way: There are two hidden layers, both with 10 neurons/units; one input layer with 3 input nodes for 3 different data features, and one output as we are considering only one product demand. Thus the feature matrix would extend to a size of $N\times 3$, and the demand amount is of a $N\times 1$ vector. Besides, no scaling or regularising term applied on this part of experiment.

The simulation data for demands, Weather and Promotion is randomly set. As for Holiday, the weekends were set with value 1, and weekdays are all 0. The demand observations are generated by Matlab 2017 'randi(20,3)' function, indicating that drawing data from uniform distribution with the range between 3 and 20.
\begin{table}[h]
\caption{Small Synthetic Data 1 (Training and Testing)}\label{table1}
\begin{center}{\normalsize 
\begin{tabular}{l|ccccccc||ccccccc} \hline
&\multicolumn{7}{|c||}{Training Data} & \multicolumn{7}{c}{Testing Data} \\
Demand & Mon & Tue & Wed & Thu & Fri & Sat & Sun & Mon & Tue & Wed & Thu & Fri & Sat & Sun\\ 
\hline 
Week1 & 13 & 7 & 16 & 7 & 12 & 15 & 19 & 7 & 10 & 6 & 5 & 18 & 12 & 18\\ 
Week2 & 20 & 12 & 5 & 5 & 7 & 18 & 7 & 17 & 19 & 7 & 5 & 13 & 5 & 14 \\ \hline 
\end{tabular} 
}\end{center}       
\end{table}

Testing error and training errors are plotted against $c_p/c_h$.

\begin{figure}[H]
 			\centering
 			\includegraphics
 			[width=5.44in,height=2.584in]
 			{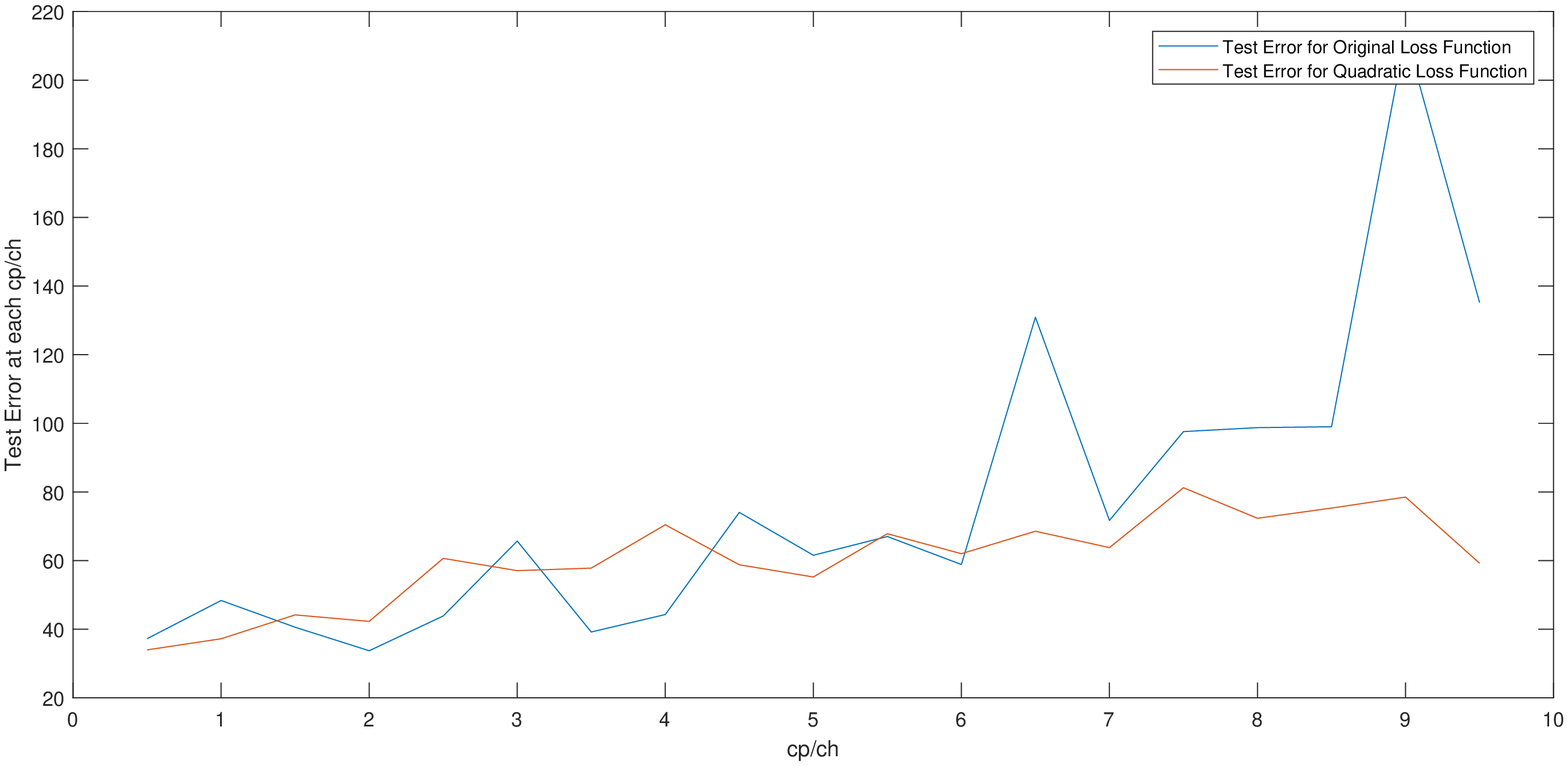}
 			\caption{Training Error Comparison For Small Synthetic Simulation}
\end{figure}  

\begin{figure}[H]
 			\centering
 			\includegraphics
 			[width=5.44in,height=2.584in]
 			{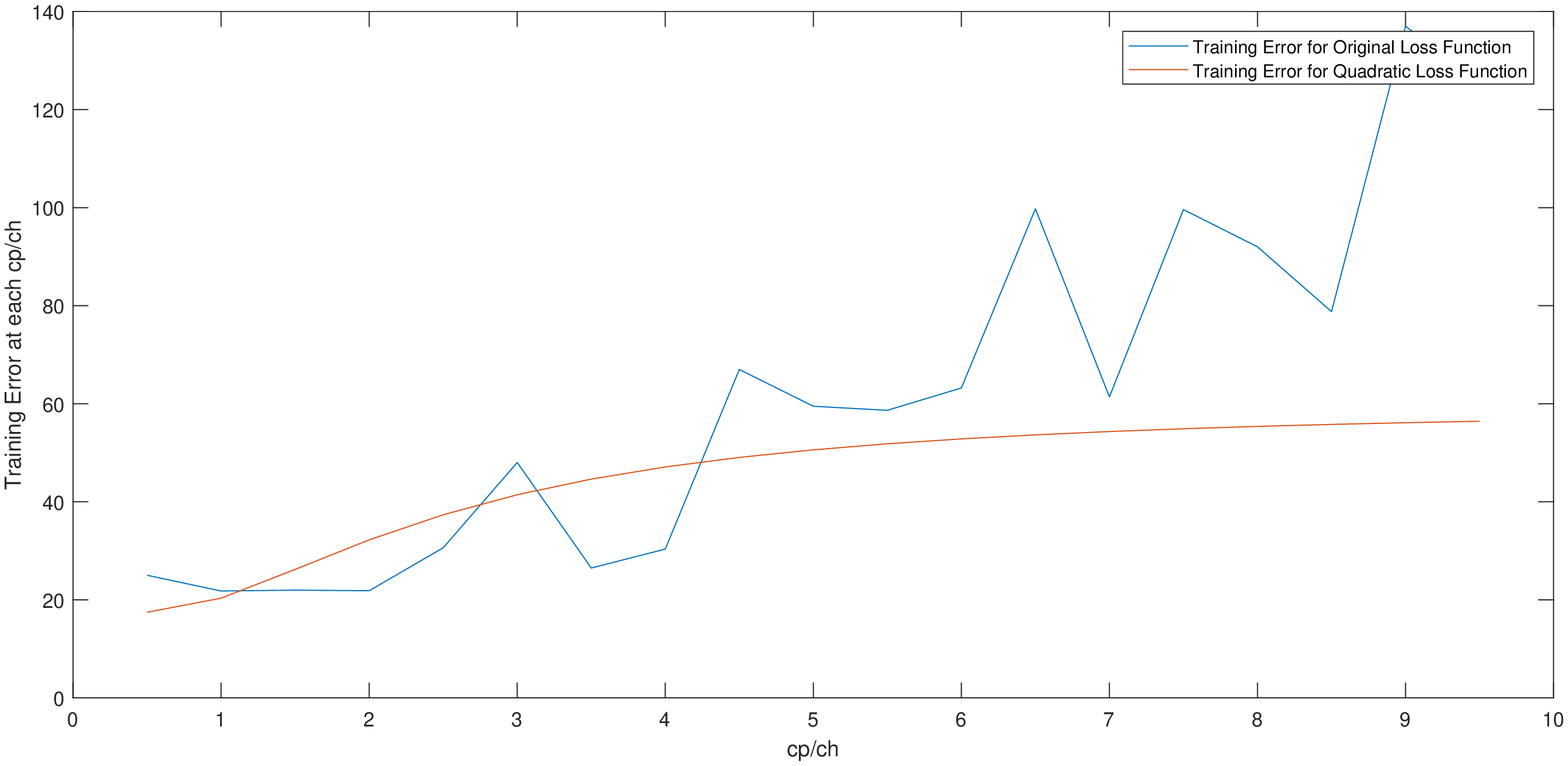}
 			\caption{Training Error Comparison For Small Synthetic Simulation}
\end{figure} 

The reason of this kind of poor performance for original loss function can be partially explained by the the nature of dataset. Here, the small synthetic data operates in a small range between 3 and 20. Lacking in large outliers in this dataset means that overfitting problem cannot be truly reflected, thus this small experiment shows both loss functions perform comparably.

\subsection{Empirical Study}
In order to make a fair comparison between the quadratic loss function and the original loss function, the real world data from a retailer \cite{Foodmart} between 1997 and 1998 is used to assess the model. This data set contains 13,170 observations for different items from 24 departments in 3 stores; 9,877 observations out of 13,170 were used for training set (around $70\%$); while the rest 3,293 observations (around $30\%$) were used as testing set. 

Here as further classification did not applied, thus this empirical dataset was still considered as a single-product Newsvendor problem. 

As the comparison between the quadratic loss function and original loss function is the main objective, the optimal DNN structure was determined based on the iterative training with Quadratic loss function, and use this structure to train the model under original loss function, which is as followed:

 		\begin{center}
 		\begin{tabular} {l*{6}{c}r} 
 			Parameters & Value \\ 
 			\hline 
 			hidden layer 1 units & 282 \\
 			hidden layer 2 units & 60 \\
 			f & 1/66\\
 			{$\lambda$} & 0.001\\
 		\end{tabular}
 		\end{center} 

Testing error and training errors are plotted against $c_p/c_h$.

\begin{figure}[H]
 			\centering
 			\includegraphics
 			[width=5.44in,height=2.584in]
 			{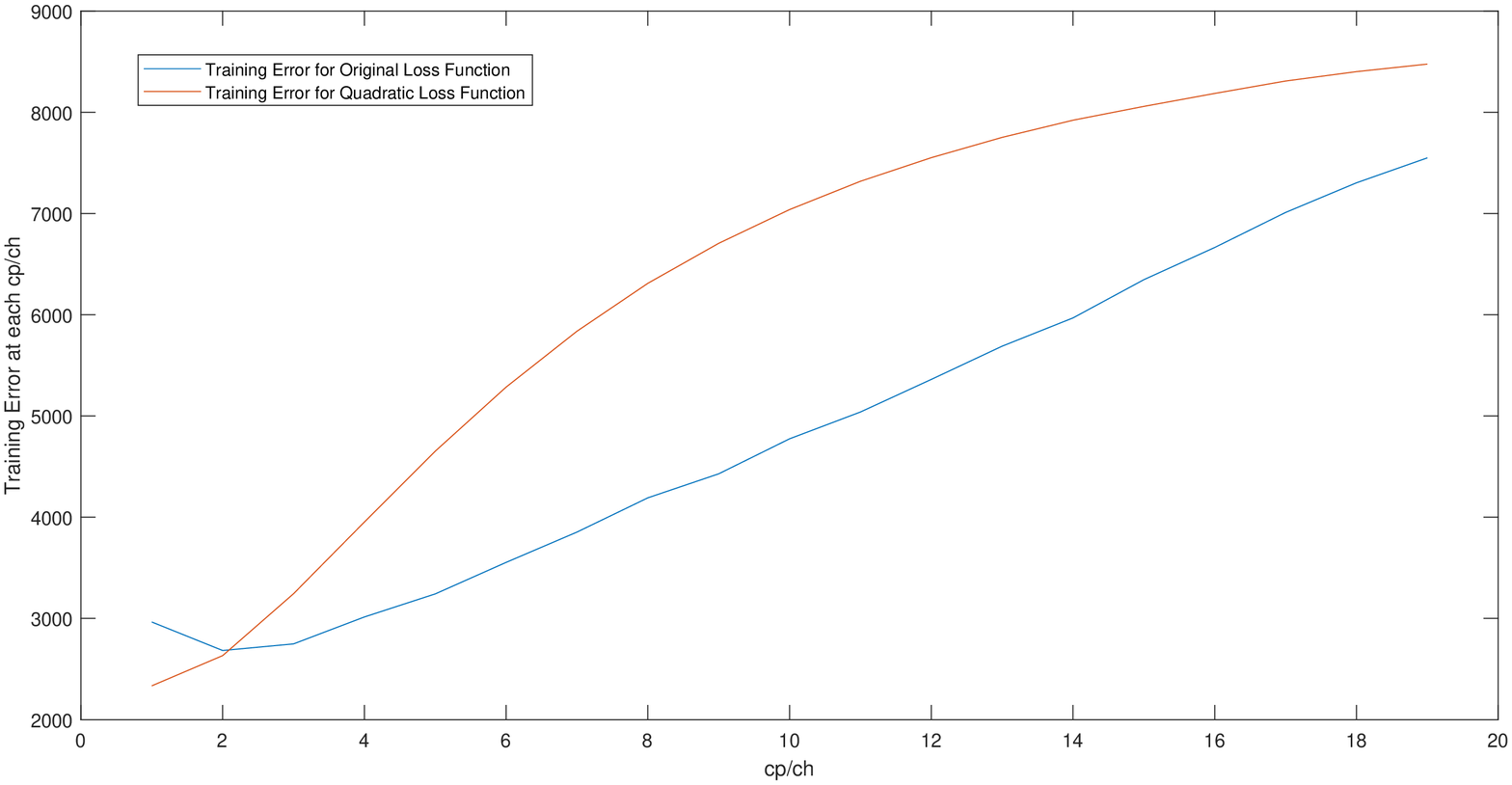}
 			\caption{Training Error Comparison For Empirical Dataset}
\end{figure}  

\begin{figure}[H]
 			\centering
 			\includegraphics
 			[width=5.44in,height=2.584in]
 			{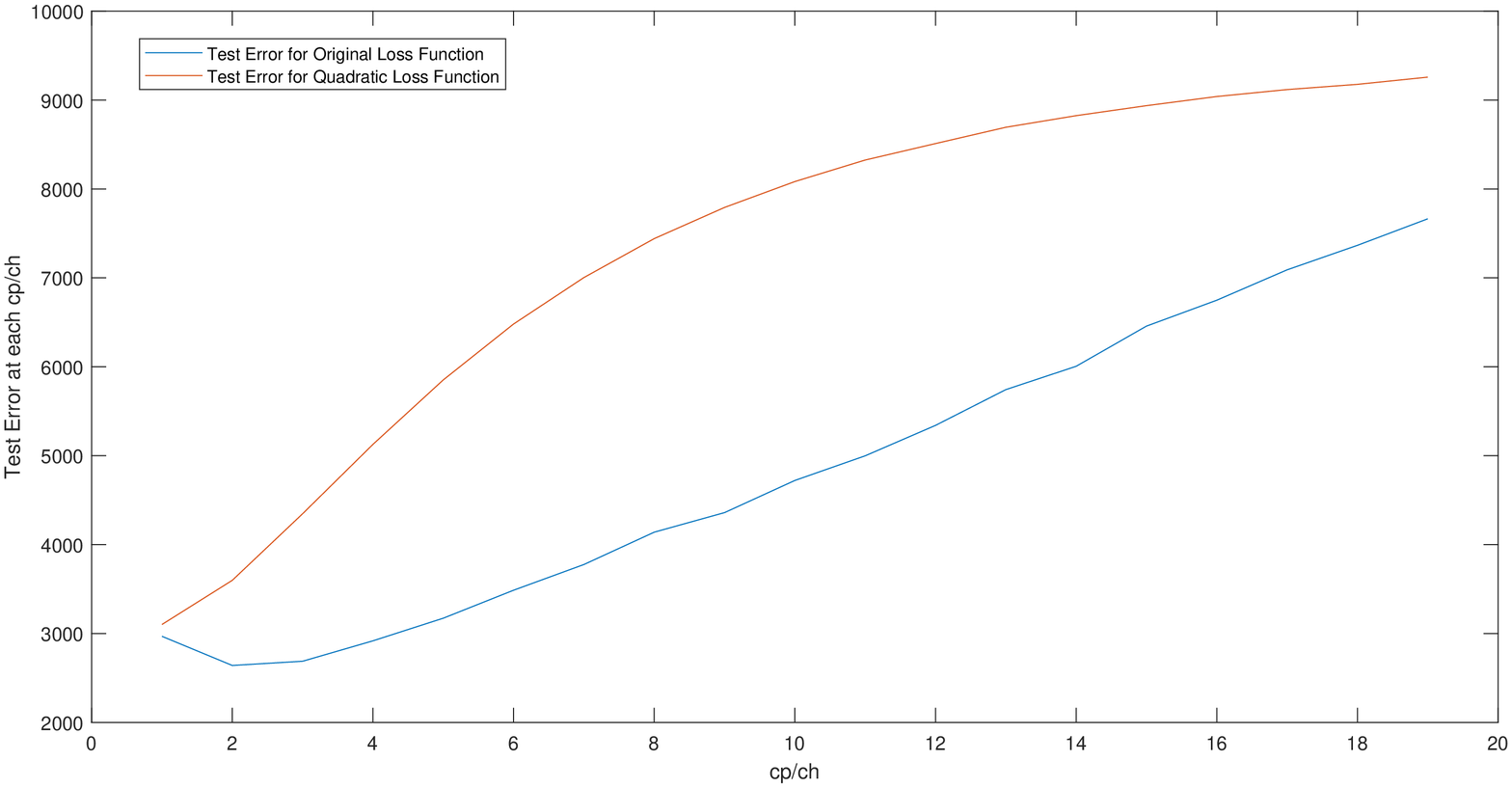}
 			\caption{Training Error Comparison For Empirical Dataset}
\end{figure} 

For further cross validating, the first 50 predicted demand in training set with the trained parameter and $c_h$=1.5, $c_p$=4 for both original and Quadratic were also attached to explain the performance:

\begin{figure}[H]
			\centering
 			\includegraphics
 			[width=5.44in,height=2.584in]
	 		{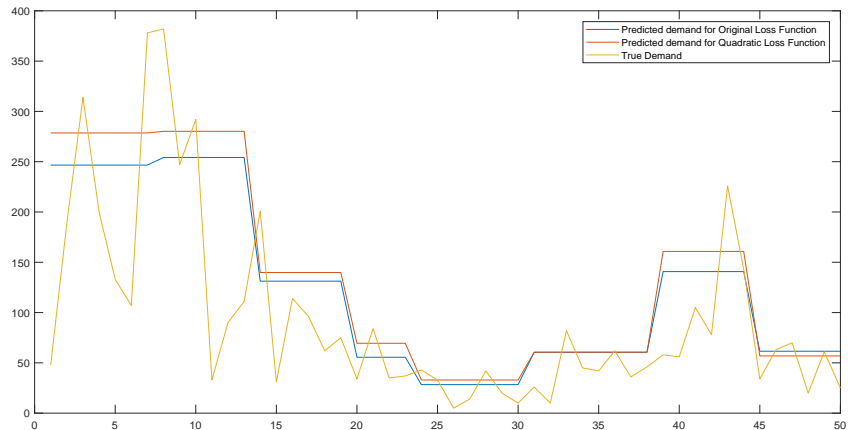}
 			\caption{Prediction Comparison in Training set For Empirical Dataset} \label{Figure4}
\end{figure} 
		
Viewing from the previous outcomes, original loss function has good performance in both in-sample and out-of-sample performance in this empirical dataset. The projected first 50 predicted demand in training set shows that original and Quadratic loss function are both good at capturing the change of demand data, and the Quadratic loss function in general generate larger values when facing with large demand observations, which further reflect the the nature of potential overfitting in Quadratic loss function.

\subsection{Testing Robustness to Demand Outliers}
Although in empirical dataset, original loss function has good performance in both in-sample and out-of-sample environment, we still want to view how, and by how much original loss function can prevent overfitting in dataset with extreme properties. Thus in this part, we view the performance of the dataset with specific properties.

\subsubsection{Simulation 1: Same Data Features With Different Demand}
When the given data is split into independent variables and dependent variable (the variable that we want to make some prediction), it is quite general to observe some observations with same values on independent variables but have different values on dependent variables. As in the fitted model, a specific set of values for independent variables should only fit with one value of dependent variable in the fitted model, if such kind of situation should happen, the fitted value of dependent variable would be largely determined by the fitting method and the observed dependent variables in the dataset.

Putting this scenario into the Newsvendor problem with data features, the problem can be transformed to: same data features with different demand. Thus, a dataset that fully consisted by such kind of data resampling from the empirical dataset can assess the performance of Quadratic loss function and original loss function, especially the problem of overfitting.

The data generating process is under following steps: first, find all the observations in the data set that shares same data features, forming those observations with same data features to be together, and calling it as 'blocks', the data has a total of 1,848 blocks. Then randomly draw a total number of 500 blocks to form the final dataset. The final dataset is with 3,565 observations. 

Testing error and training errors are plotted against $c_p/c_h$.

		\begin{figure}[H]
			\centering
			\includegraphics
			[width=5.44in,height=2.584in]
			{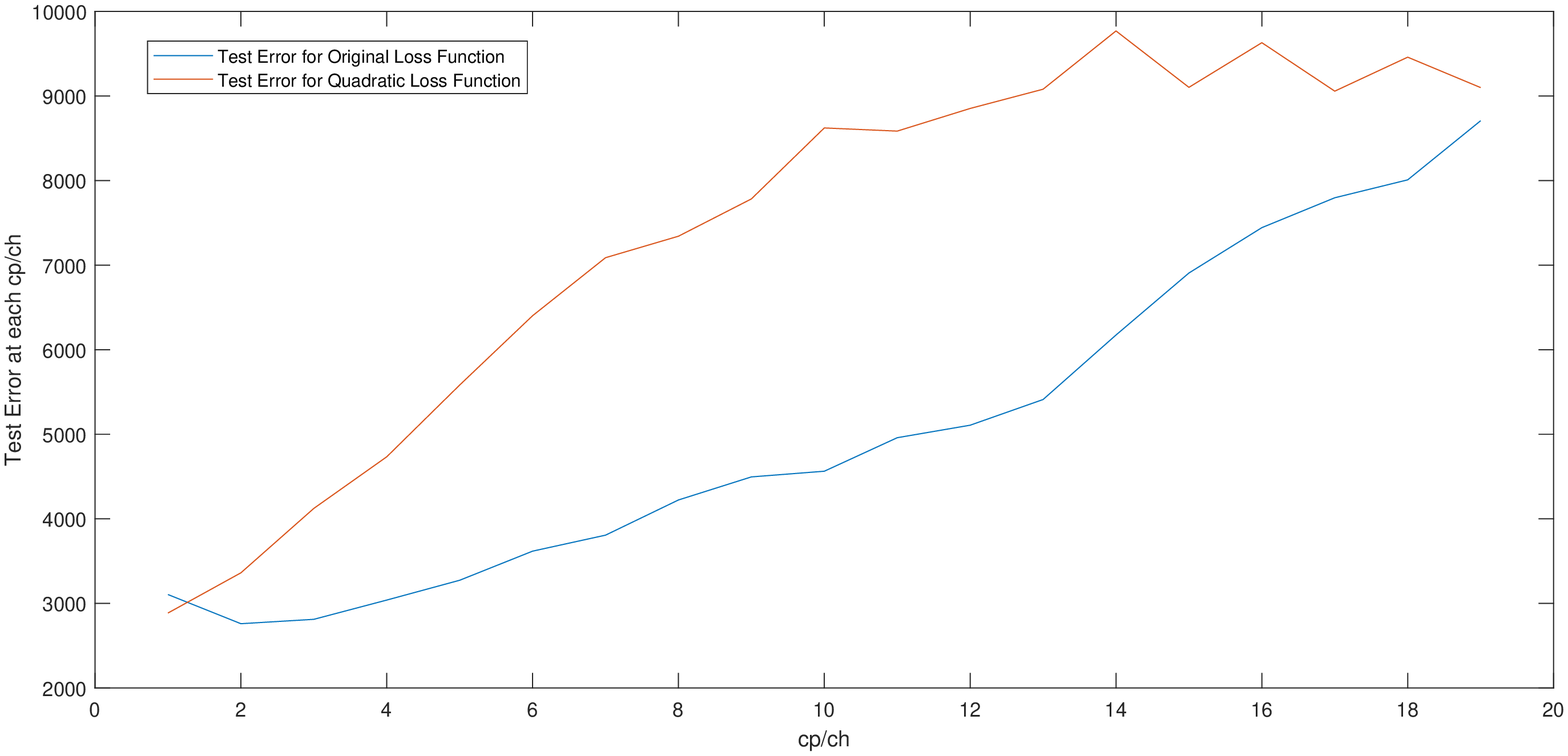}
			\caption{Testing Error Comparison For Simulation1}
		\end{figure}
        
		\begin{figure}[H]
			\centering
			\includegraphics
			[width=5.44in,height=2.584in]
			{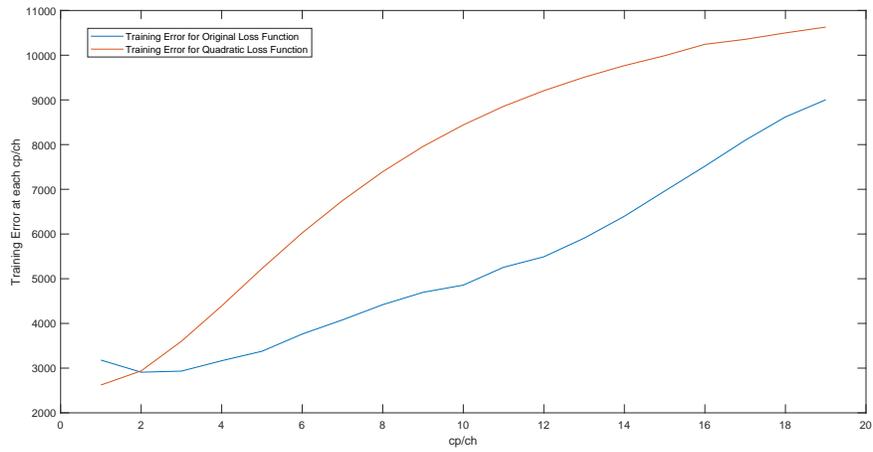}
			\caption{Training Error Comparison For Simulation1}\label{Fig8}
		\end{figure}  

For further cross validating, the first 50 predicted demand in training set with the trained parameter and $c_h$=1.5, $c_p$=4 for both original and Quadratic were also attached to explain the performance:

		\begin{figure}[H]
			\centering
			\includegraphics
			[width=5.44in,height=2.584in]
			{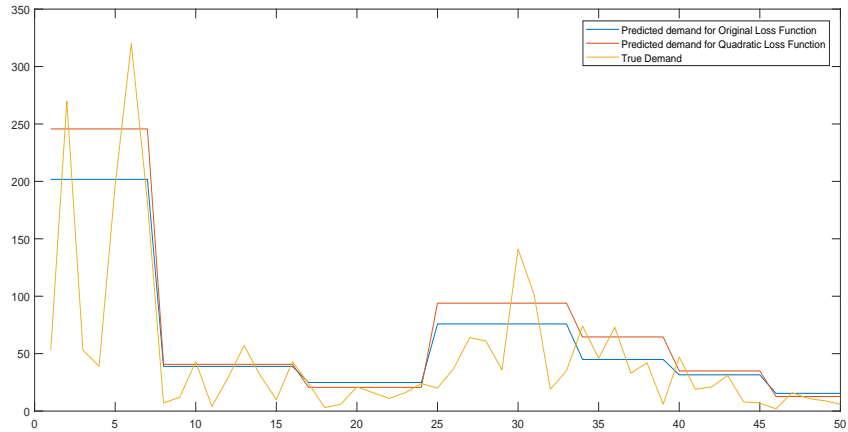}
			\caption{Prediction Comparison in Training set For Simulation1}
			\label{Fig9}
		\end{figure} 

The testing error and training error for this part is quite similar to the previous empirical study, indicating both good in-sample and out-of-sample fitness. In Figure \eqref{Fig9}, the clustering of predictions is quite obvious, which is under the form of plateaus in the predicted value, indicating that the underlying data features of these observations are the same. Here both loss function exaggerated the prediction for the first 8 observations, and original loss functions just outperforms Quadratic loss functions by a small extent, but the performance of original loss function for the first 8 observations and from observations 25 to 34 is quite good.

\subsubsection{Simulation2: Large outliers in Demand}
In the previous part, cases when several observations shares same data features but with different demand were viewed. Judging from the Figure \ref{Fig9}, the original demand data fluctuates so wildly that the prediction is hardly fit with them. This kind of characteristic can be viewed as outliers. In this part, the problem of outliers in the dataset would be exaggerated to measure the ability of avoiding outliers for both loss functions.

In this experiment, we wish to test the robustness of both loss functions against outliers in demand data. It is typical that the practical demand data can fluctuate wildly, thus the prediction is hardly fitting with them. This kind of characteristic can be viewed as outliers. In this part, the problem of outliers in the dataset would be exaggerated to measure the ability of avoiding outliers for both loss functions.     
 
We generate data as follows: First, randomly draw 1,000 examples from the empirical dataset. Second, randomly select observations with demand greater than 60, and multiply them by 10 to simulate outliers. The reason why 10 was applied for scaling was to make sure that the data exceed the largest value of the original empirical data, make them real outliers. We identify 142 outliers in transformed data. 

\begin{figure}[H]
	\centering
	\includegraphics
	[width=5.44in,height=2.584in]
	{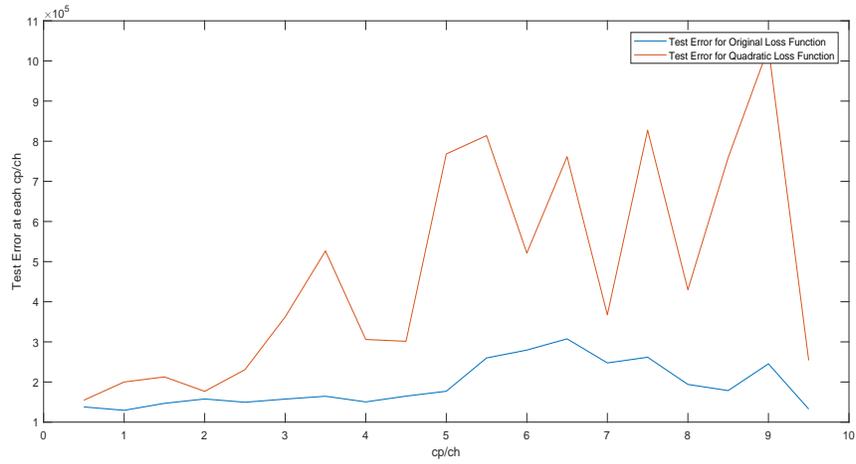}
	\caption{Testing Error Comparison For Simulation2}\label{Fig8}
\end{figure} 

\begin{figure}[H]
 			\centering
 			\includegraphics
 			[width=5.44in,height=2.584in]
 			{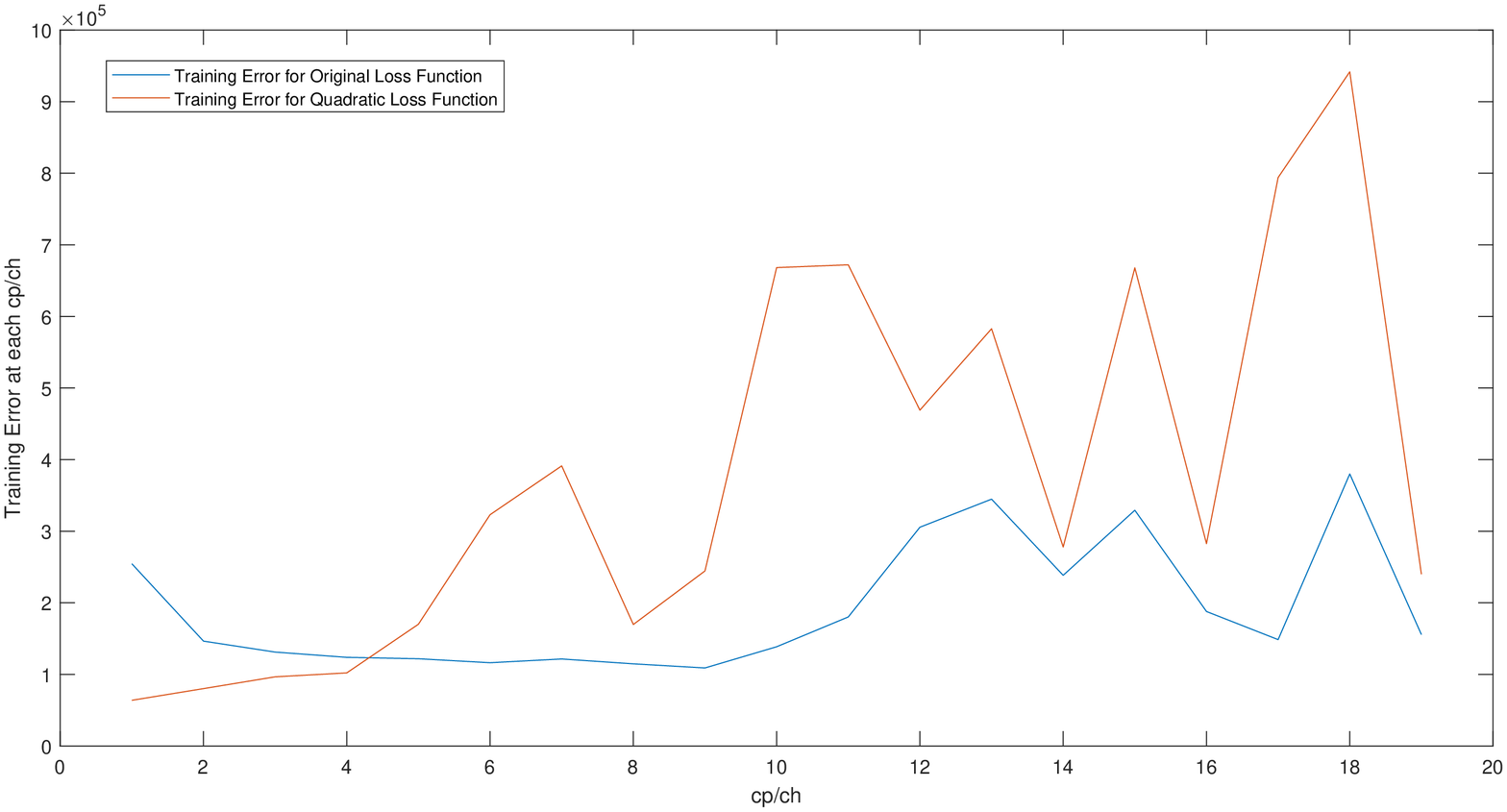}
 			\caption{Training Error Comparison For Simulation2}\label{Fig8}
\end{figure}  

For further cross validating, the first 50 predicted demand in training set with the trained parameter and $c_h$=1.5, $c_p$=4 for both original and Quadratic were also attached to explain the performance:

\begin{figure}[H]
 			\centering
 			\includegraphics
 			[width=5.44in,height=2.584in]
 			{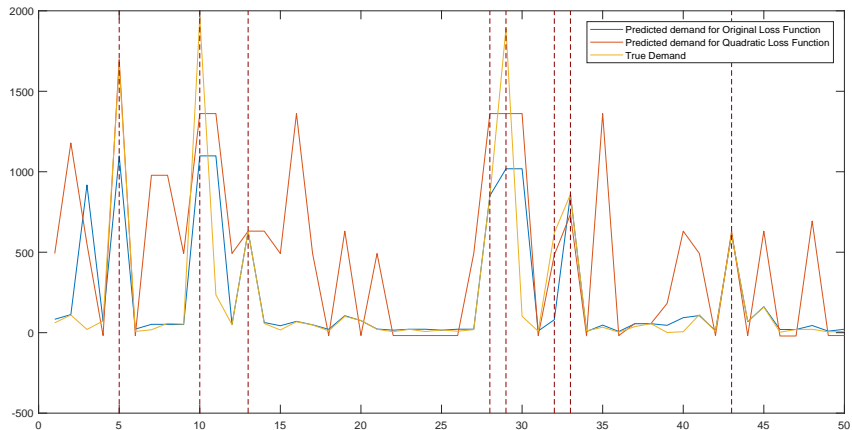}
 			\caption{Prediction Comparison in Training set}
 			\label{Figure5}
\end{figure} 

In this part, the comparison between predicted demand and true demand is the most important. The true demands with red dash lines indicating the simulated outliers generated. In general, the original loss function performs well at each outliers by not overfitting them, while the prediction given by the Quadratic loss function tries to approximate those outliers, generating a wrong fitness. Besides those outliers, the original loss function also fits the demand data well enough. Training Error for the original loss function before 5 in $c_p/c_h$ is larger than the quadratic loss function, which further explain that Quadratic loss function overfits the training set.

\section{Conclusions}\label{Sec:5}
This paper considers a new approach for the multi-feature Newsvendor problems. Several approaches in solving Newsvendor and Newsvendor-like problems were summarised. 

Most of the historical approaches solved the optimal demand distribution instead of solving the optimal demand amount directly, and those approaches that solved the demand directly did not consider large volume of historical data. The one \citep{OroojlooyjadidSnyderTakac2016} uses the deep-learning method in solving the optimal demand under large volume of historical data uses a inappropriate method \eqref{Eq4}.

We have demonstrated that the original loss function is more appropriate in solving the multi-product and multi-feature Newsvendor problems by designing a deep learning algorithm and testing for both synthetic and real-world demand data. Our experiments showed the advantage of original loss function to the quadratic loss function used in a recent research. The advantages mainly come from the ability to prevent overfitting with good in-sample fitness, which has  not been considered before. We recommend the deep learning for newsvendor problems should be trained with the original loss function for better performance. But it still needs to be noticed that for demands in a small range, this advantage of preventing overfitting would be overshadowed, which can be clearly stated in the Small Synthetic Simulation.

We would also like to point out that this method still has limitation. To the best of our knowledge, all the previous works on solving Multi-product Newsvendor model have not yet considered the relationship between different products, and how their relationship would expand along with the time. Incorporating these factors would make great improvement on the predictability.
\bibliographystyle{apacite}
\bibliography{YanfeiZ}
\end{document}